\documentclass[nonacm,sigconf]{acmart}

\AtBeginDocument{%
  \providecommand\BibTeX{{%
    \normalfont B\kern-0.5em{\scshape i\kern-0.25em b}\kern-0.8em\TeX}}}

\begin{document}

\title{An Interpretable Deep Learning System for Automatically Scoring Request for Proposals }

\author{Subhadip Maji}
\affiliation{%
  \institution{Optum, UnitedHealth Group}
  \city{Bengaluru}
  \country{India}}
\email{maji.subhadip@optum.com}

\author{Anudeep Srivatsav Appe}
\affiliation{%
  \institution{Optum, UnitedHealth Group}
  \city{Hyderabad}
  \country{India}}
\email{anudeep.srivatsav@optum.com}

\author{Raghav Bali}
\affiliation{%
  \institution{Optum, UnitedHealth Group}
  \city{Bengaluru}
  \country{India}}
  \email{raghavbali@optum.com}

\author{Veera Raghavendra Chikka}
\affiliation{%
  \institution{Optum, UnitedHealth Group}
  \city{Hyderabad}
  \country{India}}
\email{veera_raghavendra@optum.com}

\author{Arijit Ghosh Chowdhury}
\affiliation{%
  \institution{Optum, UnitedHealth Group}
  \city{Bengaluru}
  \country{India}}
  \email{arijit.ghosh@optum.com}

\author{Vamsi M Bhandaru}
\affiliation{%
  \institution{Optum, UnitedHealth Group}
  \city{Hyderabad}
  \country{India}}
  \email{vamsi.bhandaru@optum.com}

\renewcommand{\shortauthors}{Maji and Appe, et al.}

\begin{abstract}
The Managed Care system within Medicaid (US Healthcare) uses Request For Proposals (RFP) to award contracts for various healthcare and related services. RFP responses are very detailed documents (hundreds of pages) submitted by competing organisations to win contracts. Subject matter expertise and domain knowledge play an important role in preparing RFP responses along with analysis of historical submissions. Automated analysis of these responses through Natural Language Processing (NLP)  systems can reduce time and effort needed to explore historical responses, and assisting in writing better responses.  Our work draws parallels between scoring RFPs and essay scoring models, while highlighting new challenges and the need for interpretability. Typical scoring models focus on word level impacts to grade essays and other short write-ups. We propose a novel Bi-LSTM based regression model, and provide deeper insight into phrases which latently impact scoring of responses. We contend the merits of our proposed methodology using extensive quantitative experiments. We also qualitatively asses the impact of important phrases using human evaluators. Finally, we introduce a novel problem statement that can be used to further improve the state of the art in NLP based automatic scoring systems. 

\end{abstract}




\maketitle

\section{Introduction}

\textbf{Context and Scope:}
The US healthcare system is a complex setup governed and managed by state and federal agencies. Managed Care is a health delivery system utilised by Medicaid to manage cost, utilization and quality of healthcare. The Managed Care system uses contract agreements between Medicaid agencies and Managed Care Organisations (MCOs) for providing these services. Some states even utilize this system beyond traditional managed care for initiatives such as care improvement for chronic \& complex conditions, payment initiatives, etc. Contracts run the gamut from computer support to janitorial services to direct client services. HHS posts all notifications of new \textbf{Request for Proposal (RFP)}/solicitation releases, Requests for Application and Open Enrolments. RFPs are bid requests consisting of functional and non-functional requirements for different services. These also outline model contracts and the expected format of the proposals. The requirements are mentioned in the form of different questions/queries which are answered by each proposal/response to these RFPs. The procurement of these contracts entirely depends upon the scores obtained for each response based on the predefined evaluation criteria. A contract is generally awarded to the best scoring respondent(s).

A typical RFP bid consists of RFP advertisement, RFP itself, a model contract, proposals/responses from bidding entities (such as MCOs) and scoring sheets for all the submissions. RFPs and supporting documents are publicly available information. MCOs typically utilise historical submissions to understand the requirements and respond better to improve their chances of winning a bid. Every RFP response (and related documents) typically runs into several hundred pages which are spread across different websites and data stores. Manual exploration of historical bids is a time consuming and iterative process. Given the changing healthcare landscape, limited time-frame and resources to draft new responses, the current process is not comprehensive enough to extract insights and derive competitive advantage.

\textbf{Challenges:}
Apart from being an industry specific problem statement, our work also poses a unique challenge of scoring entire documents. Most relevant efforts towards automatic scoring have dealt with with short answers \cite{leacock2003c,ramachandran2015identifying} and essays \cite{attali2006automated,taghipour2016neural}. Our work deals with much larger sequence lengths, and a larger feature space to capture. Another difference with relevant literature is that RFPs are written by experts over multiple iterations, as opposed to students writing essays for evaluation. As such, this removes the need to check for superficial grammatical errors. Instead, there is a need to identify which aspects of the text enhance scores (Enablers) and those which diminish it (Disablers).

\textbf{Our Solution: }
In this paper, we propose an automated framework using interpretable natural language processing techniques to analyse RFP responses. The framework comprises of two components: Text Processing Module and an Interpretable Scoring Model. 
RFP responses usually do not follow any standard template/formatting and are available in Portable Document Format or PDF for short. Moreover, to understand the content and extract insights, the text needs to be extracted at the most granular level (usually section or question level). These issues complicate the text extraction process  and thus the need to develop a Text Processing Module. We have developed a generic Text processing module that would extract text from different formats of response. The extracted text is then analysed using our Interpretable Scoring Model. The scoring model enables us to identify terms/phrases and other auxiliary features which impact the section/question score positively and negatively. We term positively impacting features as enablers and negatively impacting ones as disablers. The framework also provides insights about auxiliary features which latently impact overall scoring. The framework also provides a single portal/platform to access historical bid responses for similar details across bidders and states. 

Major Contribution of this work is as follows, we have:
\begin{itemize}

\item 	Built a generic pdf parser to extract section-level content from RFP pdf documents. 
\item		Generation of a real world document level scoring dataset that can be used to further NLP research.
\item 	Proposed an interpretable deep-learning based regression model to automatically score RFP documents.
\item 	Addressed a novel problem of identifying enablers and disablers from RFP documents for effective writing purposes.

\end{itemize}

\section{Related Work}

Even though there are research studies on effective management of RFP pipeline, processing RFP data is a least addressed problem. USPTO: US6356909B1 presented a web-based system for management of RFP process. The system handles end-to-end pipeline of tasks such as, to generate RFP forms, the process of responding for RFPs and the process of reviewing and presenting the results. The RFPs are widely used in contract-based software development projects. Saito et al. \cite{saito} proposed a simple evaluation model to check whether the user requirements of software project were accurately mentioned in an RFP. But, the model is mainly focused on non-functional requirements like response time and security issues in the RFP. 

This work is similar to Automated Essay Scoring (ASE) when it comes to evaluating a draft. Various works \cite{crossley2018assessing,inproceedings} proposed deep learning approaches using CNNs and LSTMS to predict the grade/score of the text in Essay Grading problems. The most prominent dataset for Automated Essay scoring is the Automated Student Assessment Prize (ASAP) dataset \footnote{https://www.kaggle.com/c/asap-aes}. AES systems using the ASAP dataset often model as a regression problem \cite{kumar2019get,goenka2020esas} and then convert it into categorical variables, due to the narrow range of scoring values within the dataset.  Our dataset has much larger sequence length compared to essays, as well as a wider scoring range (0-100), which makes it a more challenging problem to solve.

Identifying positively/negatively contributing phrases is analogous to Sentiment Analysis. However, tools on sentiment analysis either use predefined dictionaries of positive words such as \textit{good, better}, etc,. and negative words such as \textit{not, worse} for polarity detection \cite{feldman2013techniques} or learn language semantics to understand subjectivity for polarity detection using huge corpora. Instead, in this paper we are generating such dictionaries with positively and negatively impacting words/phrases from RFP responses which are usually devoid of subjectivity.  In some cases, essays with inclusion of terms that are specific to the domain would result in better score \cite{burstein2005advanced}. However, they do not consider negatively contributing terms.  

We perform thorough quantitative experiments to model the scoring system, as well as qualitatively assess enablers and disablers using expert human evaluators.

\section{Dataset}
For the purpose of this work we prepared a dataset consisting of 1300 RFP responses spread across multiple years and states with mean word count per response being 5k words. The text processing module on pool of PDFs of RFP Response and Scoring documents has two components; 

\begin{itemize}
	\item  Text Extraction from PDF Documents 
	\item  Processing and Storage of the extracted text
\end{itemize}

\begin{figure*}[]
  \begin{center}
  \includegraphics[width=0.9\linewidth]{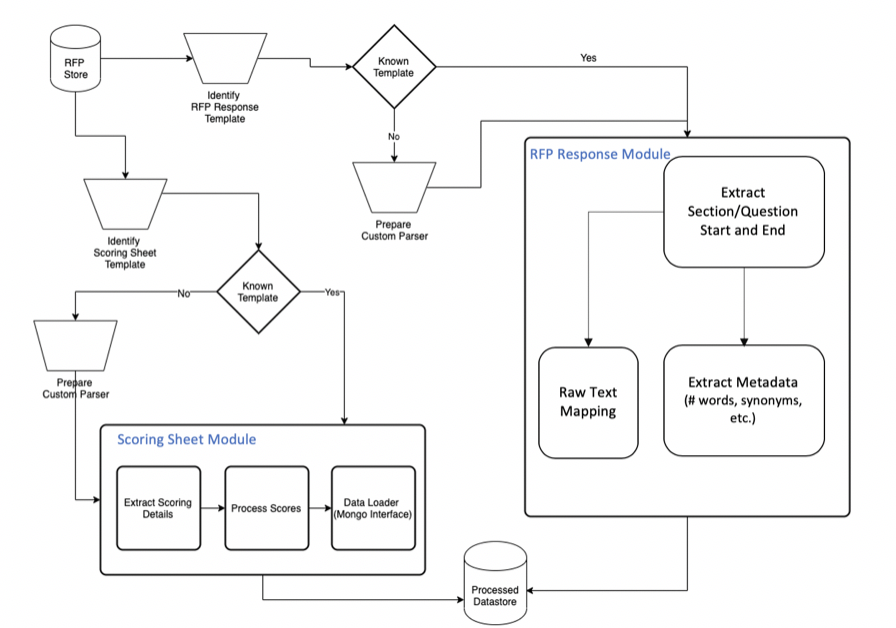}
  \caption{Text Processing Flow }
  \label{fig:text_processing_flow}
   \end{center}
\end{figure*}
\subsection{Collection}

\subsection{Text Extraction:} 
Text Extraction is the first and most important part of the overall pipeline. This step takes PDF documents as input and generates extracted text and auxiliary features as output. 
PDF is a versatile format which can handle a variety of inputs. A PDF document can be generated using text processing solutions like \LaTeX , MS Word, etc or could originate from a scan, fax or images. Usually the PDFs from text processing solutions maintain text and other attributes in native form while other sources lead to the document being an image based replica. Thus, we can broadly segregate PDFs into two categories:  
\begin{itemize}
	\item	Text-Based PDFs
	\item	Image-Based PDFs
\end{itemize}
 
Depending upon the type of PDF (text or image based), appropriate PDF parsing technique is applied. For text-based PDFs (or searchable PDFs) XML stream parsing works while OCR based methods are required for image-based PDF documents. 

For searchable PDF documents, we explored and used several XML stream parsing libraries such as pdfminer \footnote{https://www.researchgate.net/publication/267448343\_PDFMiner\_-\_Python\_PDF\_Parser}, Camelot \footnote{https://github.com/camelot-dev/camelot}. For extracting content for image-based PDF documents we utilised a proprietary OCR model.

Since there is no standardisation when it comes to RFP responses and other related documents, we performed exploratory analysis to identify common patterns across templates. This lead to creation of reusable sub-components which were used to prepare multiple custom parsers. Each parser focussed upon the following details:
\begin{itemize}
	\item	Patterns for header and footers, 
	\item	Evaluate the start and end page of particular section using headings and content tables
	\item Identify the headings and sub headings locations by leveraging the details such as font size gradients, position on the page, etc. 
\end{itemize}

The overall Text Processing and Storage flow with details related to scoring sheet module and response module is outlined in figure \ref{fig:text_processing_flow}.

\subsubsection{Scoring Sheet Module }
Every RFP bid comprises of a scoring sheet. This document usually contains tabular data with details related to different sections/questions, bidder details along with score by different evaluators. The scoring sheet module first tries to identify the template of the input document. Based on predefined rules, a specific parser is selected to extract the required information. The scoring information is normalised to maintain consistency of evaluation scale (the score ranges vary across RFP bids).

\subsubsection{RFP Response Module}
The RFP Response Module takes a PDF document as input and tries to identify if any of the existing parsers can be applied to extract information. If not, then we develop custom parsers to handle such a document. For the purposes of this paper, we developed a total 4 parsers to handle 42 templates. These 4 parsers were used for extracting text from proposals documents and scoring sheets for both searchable and scanned PDFs. Each parser was designed to extract details such as sections, question level information and a mapping of raw text with section and relevant question. We also extract details such as presence of infographics, headers, footers, tables, images, references. 

Even though RFPs and related documents are publicly available, certain portions of these documents are redacted. This is done to avoid exposure of confidential information.  We also identified percentage of redacted content per section as one of the auxiliary features.

\subsection{Processing and Storage of the Extracted Text}
The extracted metadata from Scoring sheets \& RFP Responses are cleaned and processed using standard text processing techniques. White noise in text and non-dictionary words(due to OCR conversion issues or otherwise) are removed. The following preprocessing steps we applied to the extracted text:
\begin{itemize}

	\item Removal of extra white spaces, newlines, stop words and special characters
	\item Sections, tables, URLs, emails, and date-time characters were masked using regular expressions.
\end{itemize}

Various auxiliary features derived from infographics such as number of figures/graphics, number of tables, number of words per response etc. These features are the cleaned and processed to prepare them to be used as auxiliary features for downstream tasks. We also derived additional features such as \textit{average word length, parts-of-speech tags, MCO details, degree of lexical richness, percentage of of redaction, etc.}  These auxiliary features are explained in detail in Section 5.0.1.

Final dataset along with scores, actual response texts and derived auxiliary features is stored in relevant databases. Since we had a non-standard schema for the extracted information, we made use of MongoDB \footnote{https://www.mongodb.com}, a NOSQL database.

\begin{figure*}[]
  \begin{center}
  \includegraphics[width=0.9\linewidth]{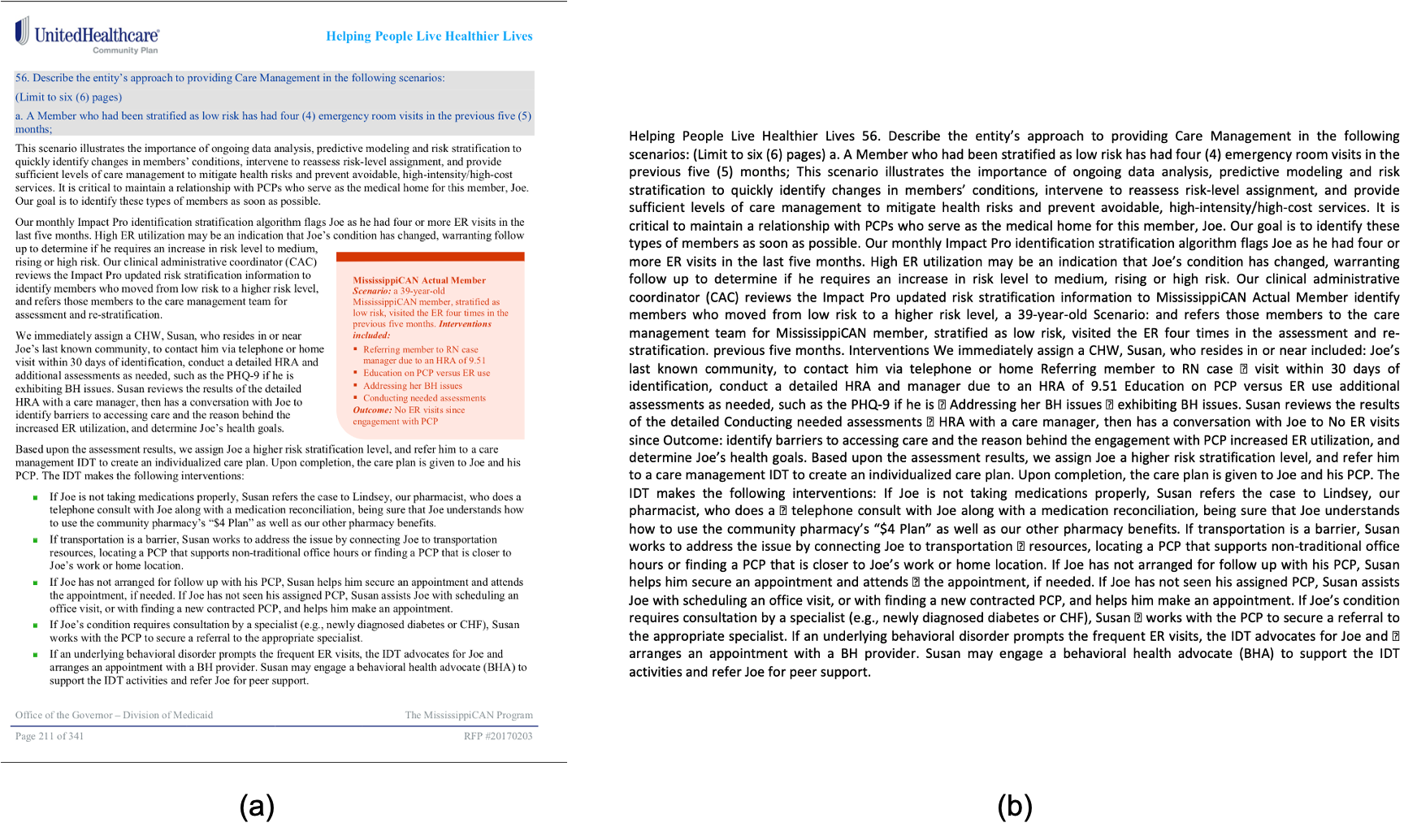}
  \caption{(a) A sample page from one of the submitted responses, (b) is the extracted text from the given page}
  \label{fig:sample_doc}
   \end{center}
\end{figure*}

\begin{figure*}[]
  \begin{center}
  \includegraphics[width=0.9\linewidth]{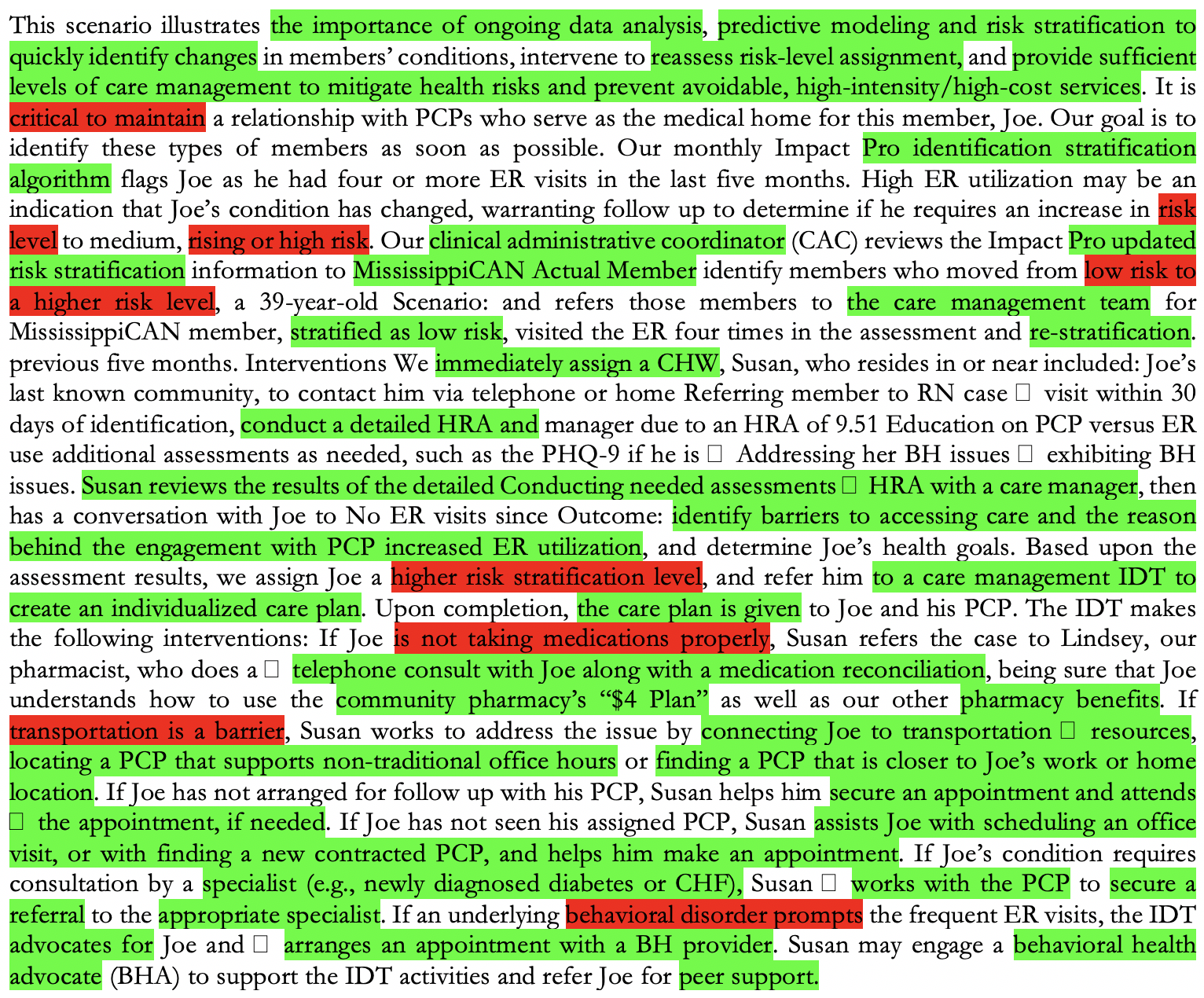}
  \caption{ Enablers and Disablers using fine-tuned Bi-LSTM model with Exclusion-Inclusion Method}
  \label{fig:ei_enabler}
   \end{center}
\end{figure*}

\section{Problem Statement}

Given a Dataset D with a questions answer pair Q  $\rightarrow$ A with a one-to-one mapping between every question and answer. A score S is assigned to every question answer pair. The goal is to predict the score based on features extracted from text (of the answer/response) and other auxiliary information. We treat this as a regression problem, since we have a high scoring range (0-100). 

\section{Model Building}
We experimented with traditional machine learning based models along with deep learning based more sophisticated approaches. This was done to ensure a comprehensive study of how different models and features impact the final setup. We also performed different ablation experiments to better understand the impact. For both the approaches, we used a set of auxiliary features apart from textual features. A brief overview of the auxiliary features is as follows:

\subsubsection{Auxiliary Features}

\begin{itemize}

\item No of words : Several research projects have shown that higher-rated essays, in general, contain more words \cite{carlson1985relationship,ferris1994lexical}. 

\item Domain ID:  Domain IDs are the broad level categorization of the responses. One response can have multiple domains. With subject matter expertise and background knowledge we created 10 domain IDs e.g. Quality Management, Compliance, Technology etc. While fitting in the model as auxiliary features we fed this domain ID feature doing one hot encoding.

\item Part of speech tags: POS tags based n-grams capture context very well. Each response word was tagged with its corresponding part-of-speech (eg., Verb, Noun, Preposition). 

\item Average Word length :  Word length can be used to indicate the sophistication of a writer \cite{hiebert2011beyond,reppen1995variation} and studies have shown that higher rated essays tend to use longer words.

\item Lexical Richness : The lexical richness is defined as the ratio of number of unique tokens present in the text to the total number of tokens present. The writer with a larger vocabulary is generally more proficient and hence is better graded than a writer using limited vocabulary \cite{reppen1995variation}.

\item Count of Sections, Figures, URLs, emails : During preprocessing Sections, Figures, URLs and Emails were replaced with the tokens \textit{SECTION}, \textit{FIGURE}, \textit{URL}, and \textit{EMAIL}. We make use of the count for these regular expressions to capture the amount of detail present in the text.

\item Doc2Vec Features: Doc2Vec \cite{le2014distributed} features are the vector representation of a document. We extracted these 300 dimensional features from a fine-tuned pre-trained Doc2Vec model  for each of the responses we have and treated these features as one of the auxiliary variables.
\end{itemize}

\subsection{Approach 1 : Random Forest }

We trained a random forest regression model on the processed text and auxiliary features as explanatory features with normalized score as dependent variable. We used a bag-of-words approach to transform textual data into usable form. We also experimented with \textit{tf-idf} based features but could not achieve any significant improvement. We fine-tuned the model for mean absolute error and adjusted $R^2$ to achieve best possible outcomes. We utilised grid-search for hyper-parameter optimisation. We used k-fold (in this exercise 5 fold) cross validation technique to ensure that a stable model has been identified by addressing bias-variance trade off issues.

\subsection{Approach 2 : Deep Learning Based Approaches}
In approach 1, the main drawback was under-utilisation of textual features. The bag-of-words feature set has its limitation as it fails to capture context and key linguistic features. To tackle this issue, we upgraded experimented with deep learning based NLP models as well.	

\begin{figure*}[]
  \begin{center}
  \includegraphics[width=0.9\linewidth]{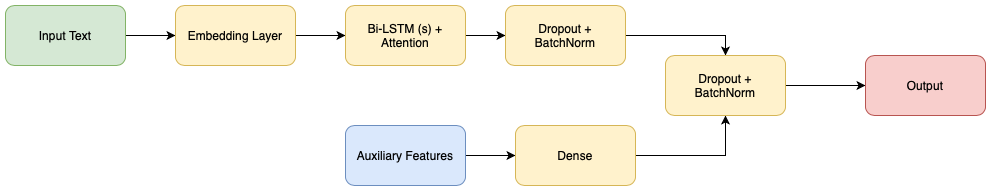}
  \caption{Bi-LSTM based scoring appoach}
  \label{fig:bilstm_model}
   \end{center}
\end{figure*}

The model setup is as follows. Textual input is handled using one or more bi-LSTM layer(s) with $b$ hidden layers $e$ dimensional embedding layer. The bi-LSTM layer is followed by a Global Average Pooling/Flatten layer, Dropout Layer with dropout rate of $r$ and Batch Normalization respectively. For auxiliary input, we put one dense layer of size $d$ and merged this with the last output of text input branch. After merging we again use one Dropout layer with rate $r$ and Batch Normalization layer. Finally one dense layer of size 1 with linear activation for the final prediction of score.
We use the different optimizers (e.g. Adam, Nadam, SGD, etc.) with learning rate $l$ and default learning decay rate. The high level model architecture is shown in figure \ref{fig:bilstm_model}.

We observed that the model's validation loss converged in about 15 epochs with default learning rate. The textual input handles variable sequence length. This is done in order to handle the large variation in length of responses. We observed a variation of sequence lengths between 900 to 0.1 million words. This ability to handle variable input lengths helps us capture complete response without the need to truncate even a single word. This is also important from the interpretability perspective, i.e. identification of enablers and disablers. The large variation in input sequence lengths restricted us to use a small batch size of 4. We also make use of attention mechanism \cite{bahdanau2014neural} to capture the distinguished influence of the words on the output prediction. 

The hyper-parameter space we tried for fitting the model is, $b \in \{32, 64, 128, 256, 512\}, e \in  \{32, 64, 128, 256, 512. 1024\}, r \in [0, 1), l \in [1e-7, 1e-2], d \in  \{32, 64, 128, 256, 512, 1024\}.$ We tried different intermediate activation functions e.g. ReLU, tanh, sigmoid, etc. We created one custom activation layer specific to our problem i.e. as our output score is always between 0 to 100, we clipped each intermediate activation to be in between 0 and 100. This custom activation function is defined as $f: z \rightarrow a$ such that, 
\[ f(z) = \begin{cases} 
0 & z\leq 0 \\
z & 0\leq z\leq 100 \\
100 & 100\leq z 
\end{cases}
\]
This is very similar to ReLU activation with an additional constraint of clipping the maximum value.
While calculating the model loss we also clipped the final prediction of the model $\hat{y}$ to be between 0 and 100, thus limiting the L1 loss value. The use of custom loss function reduced the test loss significantly. However, using either the custom loss function or the custom activation function is required as both lead to similar impact. 
Along with random initialization of embedding matrix, we tried pre-trained embeddings from BERT\cite{devlin2018bert} , GloVe \cite{pennington2014glove} and Word2vec \cite{rong2014word2vec}. 

\begin{table}[h]
		\begin{tabular}{c c c}
			\hline
			\textbf{Model Setup} & \textbf{MAE}  \\
			\hline
			\noalign{\smallskip}
			\hline 
			Random Forest & 14 \\ 
			Random Forest + Auxiliary & 12.3 \\\hline
			Bi-LSTM + Attention  & 8.7 \\
			Bi-LSTM + Attention + Auxiliary  & 8.0 \\
			Bi-LSTM + Attention + Auxiliary + BERT-Embeddings & 7.4 \\
			\hline
		\end{tabular}
		\caption{Modelling Approaches with corresponding mean absolute error}
		\label{tab:mae}     
\end{table}

\begin{center}
  \begin{table*}[h]
\begin{tabular}{c c}
\hline
\textbf{Enablers} & \textbf{Disablers}  \\
\noalign{\smallskip}
\hline 
['death', 'understand', 'social determinant',  & ['knowledgeable', 'housing', 'transportation',  \\

'leave', 'learn',  & 'department', 'utilization',  \\
'previous', 'approach', 'shelter', 'conduct', 'pharmacy'] &'peer', 'reside', 'social support', 'circumstance', 'symptom'] \\

\hline
\end{tabular}
\caption{Enablers and Disablers identified using fine-tuned Random Forest model with SHAP}
\label{tab:dataset}     
\end{table*}
\end{center}

\section{Scoring Results}

Table \ref{tab:mae} shows results for our dataset for each of the approaches explained in the previous section. The results highlight the obvious effectiveness of embeddings as compared to bag-of-words features. The deep learning approach outperforms the Random Forest model significantly, signifying that it is important to model the temporal and sequential features within the RFP responses.  

We also notice that handcrafted auxiliary features play an important role for scoring. The addition of handcrafted features improves performance within all the deep learning models, which highlights the the impact of non-textual features on the overall scoring of such documents. Our final best performing model is the Bi-LSTM model which uses BERT embeddings augmented with Auxiliary features. We would also like to highlight the effectiveness and importance of simple attention mechanisms over complex modifications, we did not observe any significant improvement using these choices. 

\section{Calculation of Enablers and Disablers}

Apart from scoring the answers, it is also important to understand which attributes contribute to a better score. This enables us to suggest better writing practices and in turn achieve higher scores.  We define \textit{Enablers} as terms which positively contribute to the score and \textit{Disablers} as terms which have a negative impact on the overall score.

\subsection{For Random Forest}

For the historical training data, out-of-bag predictions were used as bench-mark predictions. 
LIME \cite{ribeiro2016should} and SHAP \cite{lundberg2017unified}  values were extracted for each response and were rank ordered based on magnitude and direction to identify key enablers and disablers for each historical response.  Table \ref{tab:dataset} shows enablers and disablers for the sample text in Figure \ref{fig:sample_doc}(b). 

\subsection{For Deep Learning Models}

SHAP and LIME can be used for deep learning models as well. The limitation of these methods is their focus on word level importance. To provide better context and identify phrase level importance, we use the Exclusion and Inclusion (EI) method \cite{eipaper}. This method calculates the effects by  excluding phrases one at a time strategically and comparing the output score with each setting (amongst all the possible). Despite the large number of combinations this algorithm efficiently parallelises the calculations for different n-grams. The EI method has two steps for calculation of enablers and disablers for regression setting. In the first step, it calculates which words (or phrases) are not important and \textit{excludes} them by masking those words. The unimportant words are neither enablers or disablers. In the second step with only the important words remaining, it calculates the effects (either positive or negative),i.e. the \textit{inclusion} step. The effect of the words (or phrases) is evaluated through the following metric:
\begin{equation}
EI(phr_i) = \frac{\hat{y_{in}} - \hat{y_{ex}}}{\hat{y_{in}}} * 100
\end{equation}
We calculate the  percentage change in model output with respect to exclusion and inclusion of phrases. Here, $\hat{y_{in}}$ and $\hat{y_{ex}}$ are the predicted outputs including and excluding the phrase $i$ respectively. If the EI score is positive then the phrase has positive effect and if negative it has negative effect on the output. Figure \ref{fig:ei_enabler} shows the identified enabler and disabler phrases on the sample page shown in \ref{fig:sample_doc}. The phrases marked in green contribute positively towards the score (Enablers) while the phrases marked in red contribute negatively (Disablers).

\subsection{Quality of Enablers and Disablers}

Domain knowledge is essential for identification of terms that have a negative/positive impact on the document score. We performed a qualitative analysis of our results by performing a Subject Matter Experts (SME) or human evaluation of enabler and disabler terms.  We performed this exercise on a subset of 100 sample documents. From these documents, we asked the human evaluators to highlight important words and phrases ( both enablers and disablers) which are likely to help them write better answers.  Based on this, we calculate the agreement percentage of useful phrases for each document. We call this metric Phrase Quality (PQ).

  \begin{table}[h]
\begin{tabular}{c c c}
\hline
\textbf{Model} & \textbf{Method} & \textbf{PQ Agreement}  \\
\noalign{\smallskip}
\hline 
Random Forest & SHAP & 0.73 \\
Bi-LSTM & SHAP & 0.78 \\
Bi-LSTM & Exclusion-Inclusion & 0.85 \\
\hline
\end{tabular}
\caption{Phrase Quality Agreement }
\label{tab:dataset}     
\end{table}

Our method shows improved average performance over traditional methods like SHAP which only model importance at word level and do not take phrases into account. Phrases capture more context and provide better insight towards writing better RFPs.

\section{Conclusion and Future Work}

We introduced a new problem statement to NLP researchers, automatic scoring of Request for Proposals (RFP) for the insurance industry. Using a generic pdf parser, we collected data for 1300 RFP responses across multiple states in US and preprocessed it to be analysed by Natural Language Processing Pipelines. We built a scoring system using Deep Learning approaches and introduced an interpretable system for identification of enabler and disabler words and phrases. These interpretations assist experts in writing better RFP responses. Future work includes building a multimodal system that can model the aesthetic as well as content wise qualities of proposal documents.

\bibliographystyle{ACM-Reference-Format}
\bibliography{sample-sigconf}

\end{document}